\pdfoutput=1

\documentclass[11pt]{article}

\usepackage[final]{acl}

\usepackage{times}
\usepackage{latexsym}

\usepackage[T1]{fontenc}

\usepackage[utf8]{inputenc}

\usepackage{microtype}

\usepackage{inconsolata}

\usepackage{graphicx}

\usepackage{booktabs}
\usepackage{tabularx}
\usepackage{bbding}
\usepackage{arydshln}
\usepackage{colortbl}
\newcommand{\myemph}[1]{\underline{#1}}
\usepackage{multirow}
\usepackage{amsmath}
\usepackage{makecell}
\usepackage{tcolorbox}
\usepackage{CJKutf8}

\title{UAQFact: Evaluating Factual Knowledge Utilization of LLMs on Unanswerable Questions}

\author{
 \textbf{Chuanyuan Tan\textsuperscript{1}},
 \textbf{Wenbiao Shao\textsuperscript{1}},
 \textbf{Hao Xiong\textsuperscript{1}},
 \textbf{Tong Zhu\textsuperscript{1}},
\\
 \textbf{Zhenhua Liu\textsuperscript{1}},
 \textbf{Kai Shi\textsuperscript{2}},
 \textbf{Wenliang Chen\textsuperscript{1}}\thanks{Corresponding author},
\\
\\
 \textsuperscript{1}School of Computer Science and Technology, Soochow University, China \\
 \textsuperscript{2}OPPO AI Center, China
\\
 \small{
    \texttt{\{cytan17726, wbshao, hxiongxionghao, tzhu7, zhliu0106\}@stu.suda.edu.cn}, \texttt{wlchen@suda.edu.cn}
 }
\\
 \small{
    \texttt{shikai@oppo.com}
 }
}

\begin{document}
\maketitle
\begin{abstract}
Handling unanswerable questions (UAQ) is crucial for LLMs, as it helps prevent misleading responses in complex situations. While previous studies have built several datasets to assess LLMs' performance on UAQ, these datasets lack factual knowledge support, which limits the evaluation of LLMs' ability to utilize their factual knowledge when handling UAQ. To address the limitation, we introduce a new unanswerable question dataset \textbf{UAQFact}, a bilingual dataset with auxiliary factual knowledge created from a Knowledge Graph. Based on UAQFact, we further define two new tasks to measure LLMs' ability to utilize internal and external factual knowledge, respectively. Our experimental results across multiple LLM series show that UAQFact presents significant challenges, as LLMs do not consistently perform well even when they have factual knowledge stored. Additionally, we find that incorporating external knowledge may enhance performance, but LLMs still cannot make full use of the knowledge which may result in incorrect responses.~\footnote{Our code and dataset are available at~\href{https://github.com/cytan17726/UAQ_Fact}{https://github.com/cytan17726/UAQ\_Fact}}

\end{abstract}

\definecolor{mygreen}{HTML}{377e28}
\newcommand{\redcross}{\textcolor{red}{\XSolidBrush}}
\newcommand{\greentick}{\textcolor{mygreen}{\Checkmark}}

\begin{table*}[t]
\small
    \centering
    \resizebox{0.96\textwidth}{!}{
    \begin{tabular}{lccccc}
    \toprule
        \multirow{2}{*}{Dataset} & CREPE & SelfAware & FalseQA & UnknownBench & \textbf{UAQFact} \\
        & \citep{yu2022crepe} & \citep{yin-etal-2023-large} & \citep{hu-etal-2023-wont} & \citep{liu2023prudent} & \textbf{(Ours)}\\
    \midrule
        Source & Web & Web & Brainstorming & Rewrite & KG \\
        \#Questions & 8,466 & 3,369 & 4,730 & 6,323  & \textbf{13,970}\\
        \#UAQs & 2,202 & 1,032 & 2,365 & 4,251 & \textbf{6,985}\\
        \#Tasks & 1 & 1 & 1 & 1 & \textbf{3}\\
        Language & EN & EN & EN & EN & \textbf{EN \& ZH}\\
    \midrule
        Answer or Label & \greentick & \greentick & \greentick \textbf{$^*$} & \greentick & \greentick\\
        Factual Knowledge & \redcross & \redcross & \redcross & \redcross & \greentick\\
    \bottomrule
    \end{tabular}
    }

    \caption{Statistics of unanswerable question datasets (data sampled from other datasets is excluded).  \textbf{UAQFact} is a large unanswerable dataset with auxiliary \textbf{factual knowledge}. Besides, it is the only dataset that supports 3 evaluation tasks in 2 languages. \textbf{$^*$}: provide a feasible response as the answer.}
    \label{tab:compare_with_other_dataset}

\end{table*}

\section{Introduction}

Large Language Models (LLMs) have shown strong performance on a wide range of tasks, including logical reasoning and question-answering \citep{achiam2023gpt, NEURIPS2022cot, liu-etal-2024-probing, qwen}. While LLMs demonstrate remarkable performance on traditional question-answering datasets, in real-world applications, questions posed by users may not have definite or factual answers, for example: \textit{"Who is the sibling of Nero Caesar and also the father of Seti I?"}. Specifically, these questions lack factual answers since there is no supporting factual knowledge either in the real world or within the constraints of the user's context. Hence, we refer to them as \textbf{unanswerable questions} (\textbf{UAQ})\footnote{We focus on the factual questions in this paper} in this paper. When faced with UAQ, if LLMs provide counterfactual responses, they might mislead users and cause unexpected consequences. 

Several researchers have built unanswerable datasets \citep{yin-etal-2023-large,hu-etal-2023-wont,liu2023prudent} for LLMs evaluation. 
With the help of these datasets, we can effectively assess the LLMs' ability to discriminate between unanswerable questions (UAQ) and answerable questions (ABQ). However, these datasets have non-negligible shortcomings:
\textbf{(1) UAQ without explicit factual knowledge support}: The UAQs in the previous studies are mainly sourced from web-crawling \citep{yu2022crepe,yin-etal-2023-large}, brainstorming \citep{hu-etal-2023-wont}, or obtained by replacing key entities in correct sentences with fake ones \citep{liu2023prudent}. 
These existing datasets only provide answers or labels without including the supporting factual knowledge. This makes it difficult to evaluate LLMs' ability to utilize internal or external factual knowledge for handling UAQs.
\textbf{(2) English only}: To our best knowledge, the existing datasets only support evaluation in English. It might be interesting to know whether the ability can be generalized to other languages.

To overcome the above shortcomings, we introduce a new dataset UAQFact, a bilingual unanswerable question dataset in which each question is accompanied by related factual knowledge. The factual knowledge is from the Knowledge Graph in two languages, English and Chinese.
We first sample factual triples from Wikidata \citep{pellissier2016freebase}, a widely used KG, as factual knowledge. 
Then we further design question templates for different question types. Based on the factual triples and the templates, we generate UAQs and ABQs. As we possess detailed factual knowledge during the generation process, we can design Chain-of-Thought (CoT) as external knowledge for each question. 
Finally, we attach related factual triples to each question as auxiliary factual knowledge, which can be used for evaluating LLMs' ability to utilize factual knowledge in handling UAQ.

Based on UAQFact, different strategies can be employed to evaluate LLMs' ability to utilize factual knowledge when handling UAQ. In this paper, we propose three evaluation tasks: one basic task similar to traditional classification tasks in existing datasets, and two new tasks specifically designed for evaluation.
(1) \textit{Discriminating between UAQ and ABQ}: a basic task that provides UAQ and ABQ directly to LLMs, evaluating their ability to discriminate them. 
(2) \textit{Evaluating LLMs' ability to utilize \myemph{internal} factual knowledge in handling UAQ}: if related factual knowledge is stored in LLMs, can they utilize the knowledge efficiently?
(3) \textit{Evaluating LLMs' ability to utilize \myemph{external} factual knowledge in handling UAQ}: if related factual knowledge is provided for LLMs in CoT, can they utilize the clues to answer UAQ correctly?

Finally, UAQFact contains 6,985 UAQs and an equal number of ABQs, totaling 13,970 questions.
Additionally, we construct 8,686 questions for UAQs' relevant knowledge and 13,970 reasoning clues as external knowledge to support the two tasks for in-depth evaluation. All data are presented bilingually in both English and Chinese.
The statistics of relevant datasets are detailed in Table \ref{tab:compare_with_other_dataset}. From the table, we can find that UAQFact is the largest unanswerable dataset among them. Besides, with auxiliary factual knowledge and reasoning clues, UAQFact is able to support the in-depth evaluation of using factual knowledge for handling UAQ.

In summary, our contributions are as follows:

\begin{itemize}
    \item We create a new dataset, UAQFact, to evaluate LLMs' ability to handle unanswerable questions. The questions are accompanied by factual knowledge from a KG. To our best knowledge, UAQFact is the largest unanswerable dataset among the existing datasets and the first that has auxiliary factual knowledge, which makes in-depth evaluation of LLMs possible. Moreover, UAQFact is in two languages, English and Chinese.
    \item We define two new tasks to comprehensively assess LLMs' ability to utilize internal and external factual knowledge in handling UAQ, respectively. During the evaluation, we design a new metric, knowledge-aware refusal rate, to measure the performance.
    \item Based on our dataset, we evaluate across multiple series of LLMs. Insights obtained from the evaluation are summarized as follows:
    
    (1) UAQFact is a challenging benchmark for LLMs in discriminating between UAQ and ABQ.

    (2) Despite LLMs having stored extensive factual knowledge within their parameters, they fail to effectively utilize internal knowledge in this task.
    
    (3) External factual knowledge may help LLMs to discriminate UAQ and ABQ. However, LLMs still can not make full use of them in handling UAQ.
\end{itemize}

\section{Related Work}

\subsection{Unanswerable Question Datasets}

Existing unanswerable question datasets are built from multiple sources, including web-crowded \citep{yu2022crepe,yin-etal-2023-large} and brainstorming \citep{yin-etal-2023-large}. The number of UAQs in these datasets is limited because UAQs are naturally rare in the real world.
To scale up the dataset, \citet{liu2023prudent} builds a large synthetic unanswerable question dataset. It collects a list of false entities and constructs UAQs by filling false entities in predefined templates or replacing key entities in ABQs. 
These datasets focus on evaluating LLMs' ability to discriminate UAQ and ABQ in English, only providing answers or labels without including the supporting factual knowledge. This makes it difficult to evaluate LLMs' ability to apply internal or external factual knowledge for handling UAQs.
Our dataset UAQFact is created from scratch based on a multilingual Knowledge Graph, collecting triples as auxiliary factual knowledge to construct UAQs and ABQs, which makes in-depth evaluations of LLMs possible.

\subsection{Evaluation of LLMs' Internal Knowledge}

Researchers have proposed several benchmarks to evaluate LLMs' internal knowledge by open-ended generation \citep{joshi2017triviaqa,paperno2016lambada,wu2024seal,lin2021truthfulqa}. While the open-ended generation setting assesses LLMs' ability to "speak out" their internal knowledge, it isn't easy to evaluate \citep{chang2024survey}. Alternatively, multiple-choice is adopted in many benchmarks as a feasible setting, including MMLU \citep{hendryckstest2021}, C-Eval \citep{huang2024c}, and LogiQA\citep{liu2020logiqa}. Therefore, for the convenience of evaluation, following the setup of previous research, we design the questions for querying knowledge relevant to UAQ in the multiple-choice format.

\begin{figure*}[t]
    \centering
    \includegraphics[width=0.92\textwidth]{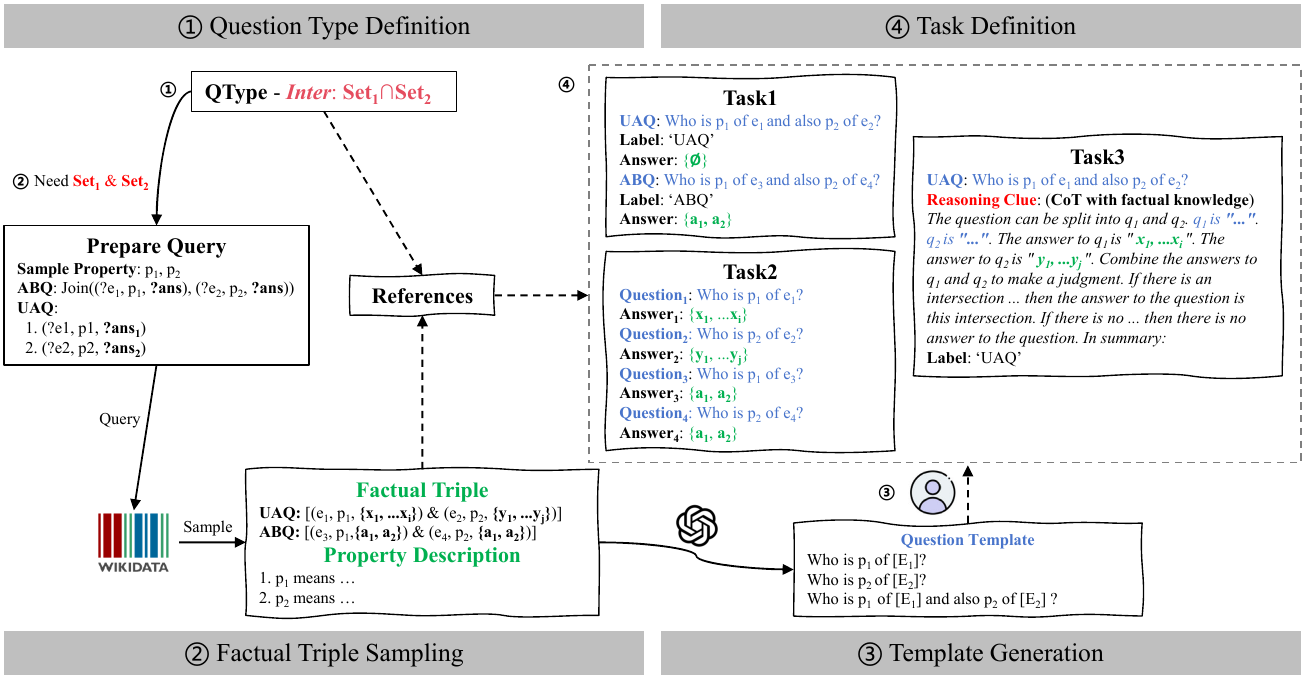}
    \caption{Dataset Construction Process (QType \textit{Inter} in English as an example) for unanswerable question (\textbf{UAQ}) and answerable question (\textbf{ABQ}): (1) Define the question type. (2) Sample factual triples from Wikidata as knowledge. (3) Generate questions by filling in the templates generated by LLM. (4) Define three tasks and compose unique inputs with factual knowledge from the preceding steps as references.}
    \label{fig_Dataset_Construction_Process}
\end{figure*}

\begin{table*}[t]
    \footnotesize
    \centering
    \resizebox{0.92\textwidth}{!}{
    \begin{tabularx}{\textwidth}{llX}
    \toprule
        QType & Description & Example\\
    \midrule
        Inter & Return intersection of two sets & (UAQ) \textbf{\textit{Q$_{i_1}$}}: \textit{Who is the \textbf{editor of Enneads} and also the \textbf{cast member in The Sixth Sense}?}\\
    \cmidrule(lr){3-3}
        & & (ABQ) \textbf{\textit{Q$_{i_2}$}}: \textit{Who is the \textbf{editor of Die Rote Fahne} and also the \textbf{cast member in The Eternal Jew}?}\\
    \midrule
        Time & Consider time constraints & (UAQ) \textbf{\textit{Q$_{t_1}$}}: \textit{Erfurt was twinned with which city \textbf{from 1957 to 1962}?}\\
    \cmidrule(lr){3-3}
        & & (ABQ) \textbf{\textit{Q$_{t_2}$}}: \textit{Erfurt was twinned with which city \textbf{from 1971 to 1976}?}\\
    \midrule
        Dilemma & Provide candidate answer & (UAQ) \textbf{\textit{Q$_{d_1}$}}: \textit{What tribe does Segestes belong to, \textbf{Mohawk people} or \textbf{Khamti people}?}\\
    \cmidrule(lr){3-3}
        & & (ABQ) \textbf{\textit{Q$_{d_2}$}}: \textit{What tribe does Segestes belong to, \textbf{Khamti people} or \textbf{Cherusci}?}\\
    \bottomrule
    \end{tabularx}
    }
    \caption{Question type (\textbf{QType}) of unanswerable question (\textbf{UAQ}) and answerable question (\textbf{ABQ}) in UAQFact. Examples in Chinese are listed in Appendix \ref{sec_app_task1_example}.}
    \label{tab:main_que_examples}
\end{table*}

\section{UAQFact}
In this section, we introduce our dataset \textbf{UAQFact}, a bilingual unanswerable question dataset in which each question is accompanied by related factual knowledge.
First, we introduce the question generation procedure (\S~\ref{sec_question_generation}) for unanswerable questions (UAQ) and answerable questions (ABQ).
Then we illustrate three tasks defined on UAQFact (\S~\ref{sec_task_definition}).
Finally, we report statistics of UAQFact (\S~\ref{sec_dataset_statistics}).

\subsection{Question Generation}
\label{sec_question_generation}

As shown in Figure \ref{fig_Dataset_Construction_Process}, the question generation has 3 steps: \textit{Question Type Definition}, \textit{Factual Triple Sampling}, and \textit{Template Generation \& Filling}. 

\paragraph{Question Type Definition} 
Inspired by widely used QA datasets~\citep{yih2016value,gu2021beyond} and relevant unanswerable question datasets~\citep{hu-etal-2023-wont}, we define three question types (QTypes): \textit{Inter}, \textit{Time}, and \textit{Dilemma}.
\textbf{(1) \textit{Inter}}: LLMs need to return the intersection of two non-empty sets, which correspond to the answer sets of two sub-questions. For UAQ, this intersection is an empty set. 
\textbf{(2) \textit{Time}}: LLMs need to respond based on the time constraints given in the question. However, such constraints cannot be satisfied for UAQ.
\textbf{(3) \textit{Dilemma}}: LLMs need to answer questions that provide candidate answers, but for UAQ, all provided candidates are incorrect.
Table \ref{tab:main_que_examples} shows the examples for each QType.

\paragraph{Factual Triple Sampling}
\label{sec_factual_triple_sampling}
Once the question type is determined, we need to acquire bilingual factual knowledge to construct the questions.
We sample factual triples as knowledge from Wikidata \citep{pellissier2016freebase}, a reliable and extensive KG that serves as a central repository for structured multilingual factual data across various subjects. 

First, we send a query to Wikidata via API \footnote{https://query.wikidata.org/} to fetch properties\footnote{Property in Wikidata can be interpreted as a relation or an attribute in triples.} and their corresponding descriptions. The property description provides the meaning and usage of the property. 
Then we define the following criteria to choose properties: 
(1) It can be easily understood with the help of its description, 
(2) It appears at least 5 times in Wikidata, and 
(3) It is capable of providing factual knowledge. 
Through the aforementioned criteria, we obtained a property list \textbf{\textit{P$_l$}}, with 724 properties, e.g. \textit{editor} and \textit{cast member}. Finally, we construct different queries corresponding to each QType to retrieve relevant entities with both English and Chinese labels. These entities combine with properties, yielding factual triples that serve as knowledge. Examples and details of factual triple sampling for three QTypes are listed in Appendix \ref{sec_app_Factual Triple Sampling}.

\paragraph{Template Generation and Filling}
\label{sec_question_generation_template}
Till now, we have factual triples and planned answers to use in the subsequent steps. 
The next step is to generate bilingual templates and fill relevant entities into the templates to generate questions.
In our approach, we first mask the entities in the relevant factual triples and keep properties unchanged, using \textit{[E$_i$]} to mask the entities intended to appear in the question and \textit{[Ans]} to mask the position intended as the question target, e.g. \textit{([E$_1$], editor, [Ans]) \& ([E$_2$], cast member, [Ans])}.
Then we provide them to GPT-3.5 along with the property description to generate templates in the target language, e.g. "\textit{editor is the person who checks and corrects a work (such as a book, ... etc.)}". The prompt we use is shown in Appendix \ref{sec_app_Template_Generation}.
The question templates generated by GPT-3.5\footnote{gpt-3.5-turbo-0125} contain errors in some cases, including semantic errors, and slots missing. Examples are listed in Table \ref{tab:tmp_Question_Generation}. To ensure the quality of the question templates, we manually inspect all templates generated by GPT-3.5, revise or discard incorrect ones, and obtain 864 templates for each language, e.g. "\textit{Who is the editor of [E$_1$] and also the cast member in [E$_2$]?}". 
Finally, we fill in the templates with entities and get the questions. 
For example, we put "\textit{Enneads}" and "\textit{The Sixth Sense}" into slots \textit{[E$_1$]} and \textit{[E$_2$]} of the template and then get the question "\textit{Who is the editor of Enneads and also the cast member in The Sixth Sense?}".

Through the above processes, we sampled factual triples from Wikidata (\textit{["(Enneads, editor, \{Porphyry, ...x$_i$\}) \& (The Sixth Sense, cast member, \{Mischa Barton, ...y$_j$\}])}") and generated corresponding ABQ or UAQ by filling in the template, e.g. "\textit{Who is the editor of Enneads and also the cast member in The Sixth Sense?"} (UAQ \textit{Q$_{i_1}$} in Table \ref{tab:main_que_examples}). The answers to these questions are decided as well. For example, "\textit{\{Rosa Luxemburg, ...\}}" for ABQ \textit{Q$_{i_2}$} and "\textit{None}" for UAQ \textit{Q$_{i_1}$} in Table \ref{tab:main_que_examples}.

\subsection{Task Definition}
\label{sec_task_definition}
In this section, we define three tasks for evaluating LLMs’ ability to utilize factual knowledge for handling UAQ and introduce their distinctive input drawing from the generated questions and factual triples outlined in \S~\ref{sec_question_generation}.

\paragraph{Task 1: Discriminating between UAQ and ABQ}
\label{sec_data_task1}
In this task, we intend to examine LLMs' ability to discriminate UAQ and ABQ. Following the settings of previous works \citep{yin-etal-2023-large,liu2023prudent,amayuelas2023knowledge}, we provide questions generated in \S~\ref{sec_question_generation} as LLMs' input and attach the corresponding answer (factual answer set to ABQ and "\textit{None}" to UAQ) for evaluation.

\paragraph{Task 2: Evaluating LLMs' ability to utilize \myemph{internal} factual knowledge in handling UAQ} 
\label{sec_data_task2}
In this task, we first probe LLM's capacity of knowledge relevant to Task 1 questions and then combine it with Task 1 results to evaluate whether LLM can effectively utilize its internal factual knowledge to handle UAQ.

We probe LLMs by asking questions about relevant factual knowledge of UAQ. If LLMs answer correctly, we consider their internal knowledge to be correct; otherwise, we consider them to be incorrect.
We construct the input in the form of multiple-choice questions, each offering four options. The following outlines the construction procedure according to QTypes. 

\begin{itemize}
    \item For a UAQ from \textbf{\textit{Inter}}, we first split the question into two sub-questions and provide four options for each sub-question. For example, \textit{Q$_{i_1}$} in Table \ref{tab:main_que_examples}, relevant triples are \textit{(Enneads, editor, ans$_1$)} and \textit{(The Sixth Sense, cast member, ans$_2$)}. We can construct two questions \textit{Q$_1$}: "\textit{Who is the editor of Enneads?}" and \textit{Q$_2$}: "\textit{Who is the cast member in The Sixth Sense?}". When constructing the options list, we ensure that both \textit{ans$_1$} and \textit{ans$_2$} appear in the option lists of the two sub-questions. Subsequently, we sample entities of the same type from Wikidata as incorrect options for \textit{Q$_1$} and \textit{Q$_2$}, thus completing their options list with four options.
    
    \item For a UAQ from \textbf{\textit{Time}}, we locate the necessary time boundary for solving it and construct the corresponding multiple-choice question. For example, \textit{Q$_{t_1}$} in Table \ref{tab:main_que_examples}, the necessary time boundary is \textit{(Erfurt, twin\_city\_start\_time, 1971)}, demonstrating that the time constraint (“\textit{from 1957 to 1962}”) is unfeasible. We can construct the following question: "\textit{When did Erfurt first twin with a city?}". Apart from the gold answer "\textit{1971}" and end time point "\textit{1962}" in the question, we randomly sample two time points, one is earlier than "\textit{1957}" and another is later than "\textit{1962}", e.g. "\textit{1954}" and "\textit{1965}" respectively.
    
    \item For a UAQ from \textbf{\textit{Dilemma}}, after eliminating candidate answers, we utilize the remaining content as the question. For example, \textit{Q$_{d_1}$} in Table \ref{tab:main_que_examples}, relevant triple is \textit{(Segestes, tribe, Cherusci)}. "\textit{Mohawk people}" and "\textit{Khamti people}" are provided candidate answers. After elimination, we obtain the question "\textit{What tribe does Segestes belong to?}". When constructing the list of options, we ensure that the gold answer "\textit{Cherusci}" and candidate answers provided in \textit{Q$_{d_1}$} are included. Subsequently, we expand the number of options to four following the entity sampling strategy outlined in \textbf{\textit{Inter}}.
\end{itemize}

\paragraph{Task 3: Evaluating LLMs' ability to utilize \myemph{external} factual knowledge in handling UAQ}
\label{sec_data_task3}

In this task, we provide LLMs with well-designed reasoning clues as external factual knowledge, evaluating LLMs' ability to utilize external factual knowledge in handling UAQ. A reasoning clue is in the form of CoT: (1) decompose input question \textit{Q} into several steps (containing question from Task 2), (2) come up with relevant factual knowledge, and (3) answer \textit{Q} based on the preceding information. In the reasoning clue, decomposed questions and relevant factual knowledge (Steps 1 \& 2) are provided.

We take \textit{Q$_{i_1}$} in Table \ref{tab:main_que_examples} as an example, the reasoning clue for it is: "\textit{\textbf{(1)} The question can be split into q$_1$ and q$_2$. q$_1$ is "Who ...". q$_2$ is "Who ...". \textbf{(2)} The answer to q$_1$ is "Porphyry, ...". The answer to q$_2$ is "Mischa Barton, ...". \textbf{(3)} Combine the answers to q$_1$ and q$_2$ to make a judgment. If there is an intersection ... then the answer to the question is this intersection. If there is no ... then there is no answer to the question.}"
More examples are shown in Appendix \ref{sec_app_Reasoning_Clues}.

\begin{table}[t]
\small
\centering
\resizebox{0.46\textwidth}{!}{
    \begin{tabular}{lc}
    \toprule
        Number of entities & 9,021 \\
        Number of properties & 724 \\
        Number of Tasks & 3 \\
    \midrule
        Number of questions (\textit{EN} | \textit{ZH}) &  \\
        $\quad$Total & 13,970\\
        $\quad$\textit{UAQ} / \textit{ABQ} & 6,985 / 6,985 \\
        $\quad$\textit{Inter} / \textit{Time} / \textit{Dilemma} & 3,428 / 3,882 / 6,660 \\

    \bottomrule
    \end{tabular}
}
    \caption{Statistics of UAQFact. We have English and Chinese versions for each question across three tasks.}
    \label{tab:dataset_statistics}
\end{table}

\subsection{Dataset Statistics and Manual Inspection}
\label{sec_dataset_statistics}

Statistics of UAQFact are presented in Table \ref{tab:dataset_statistics}. UAQFact has three distinctive features: 
First, it is a large dataset comprising 13,970 questions. 
Second, it provides auxiliary factual knowledge for each question, which can support evaluation tasks on factual knowledge application.
Third, it supports three tasks across two languages. Beyond one basic task, the other two are new tasks designed to comprehensively assess LLMs’ capabilities of using internal and external factual knowledge they have.
To ensure quality, we conducted manual inspection of UAQFact during and after its construction. The inspection result shows that 99.2\% of questions meet our standards. Details are available in Appendix \ref{sec_app_manual_inspection}.

\begin{table}[t]

\centering
\resizebox{0.48\textwidth}{!}{
\begin{tabular}{lcccccccc}
\toprule
    \textbf{Language} & \multicolumn{4}{c}{\textbf{English}} & \multicolumn{4}{c}{\textbf{Chinese}} \\
    \cmidrule(lr){2-5} \cmidrule(lr){6-9}
    \textbf{Metric} & \multicolumn{3}{c}{\textbf{Refusal Rate}} & \textbf{Acc} & \multicolumn{3}{c}{\textbf{Refusal Rate}} & \textbf{Acc}\\
    \cmidrule(lr){1-1} \cmidrule(lr){2-4} \cmidrule(lr){5-5} \cmidrule(lr){6-8} \cmidrule(lr){9-9}
    \textbf{Model} & \textbf{R$_{ua}$ $\uparrow$} & \textbf{R$_{ab}$ $\downarrow$} & \textbf{R$_\Delta$ $\uparrow$} & \textbf{Acc $\uparrow$} & \textbf{R$_{ua}$ $\uparrow$} & \textbf{R$_{ab}$ $\downarrow$} & \textbf{R$_\Delta$ $\uparrow$} & \textbf{Acc $\uparrow$}\\
    \midrule
    \multicolumn{9}{c}{\textbf{Open-sourced LLMs}}\\
    Llama3 & 38.80 & \textbf{17.91} & 20.89 & 53.21 & 23.65 & \textbf{13.70} & 9.95 & 35.88 \\
    Mistral0.2 & \textbf{62.15} & 31.27 & \textbf{30.88} & \textbf{54.32} & \textbf{49.49} & 41.57 & 7.92 & 19.31 \\
    Qwen2.5 & 59.10 & 29.03 & 30.07 & 47.73 & 48.63 & 32.00 & \textbf{16.63} & \textbf{43.77} \\
    GLM4 & 55.05 & 31.22 & 23.83 & 49.63 & 47.03 & 32.90 & 14.13 & 41.53 \\
    Average & 53.78 & 27.36 & 26.42 & 51.22 & 42.20 & 30.04 & 12.16 & 35.12 \\

    \midrule
    \multicolumn{9}{c}{\textbf{Black-box LLMs}}\\
    Gemini-1.5-pro & 66.50 & \textbf{12.25} & 54.24 & \textbf{69.62} & 57.42 & 19.01 & \textbf{38.40} & \textbf{53.98} \\
    GPT-4o-mini & 85.05 & 42.15 & 42.91 & 50.51 & 45.84 & \textbf{27.70} & 18.14 & 47.02 \\
    GPT-4 & \textbf{85.70} & 22.43 & \textbf{63.26} & 66.79 & \textbf{63.85} & 29.33 & 34.52 & 51.02 \\
    Average & 79.08 & 25.61 & 53.47 & 62.31 & 55.70 & 25.35 & 30.35 & 50.67\\
\bottomrule
\end{tabular}
}
\caption{Refusal rate and Acc of LLMs evaluated in Task 1. R$_{ua}$, R$_{ab}$: the refusal rate of UAQ and ABQ respectively. R$_\Delta$: R$_{ua}$ - R$_{ab}$}  
\label{tab_task1_result}
\end{table}

\begin{figure*}[t]
    \centering
    \includegraphics[width=0.96\textwidth]{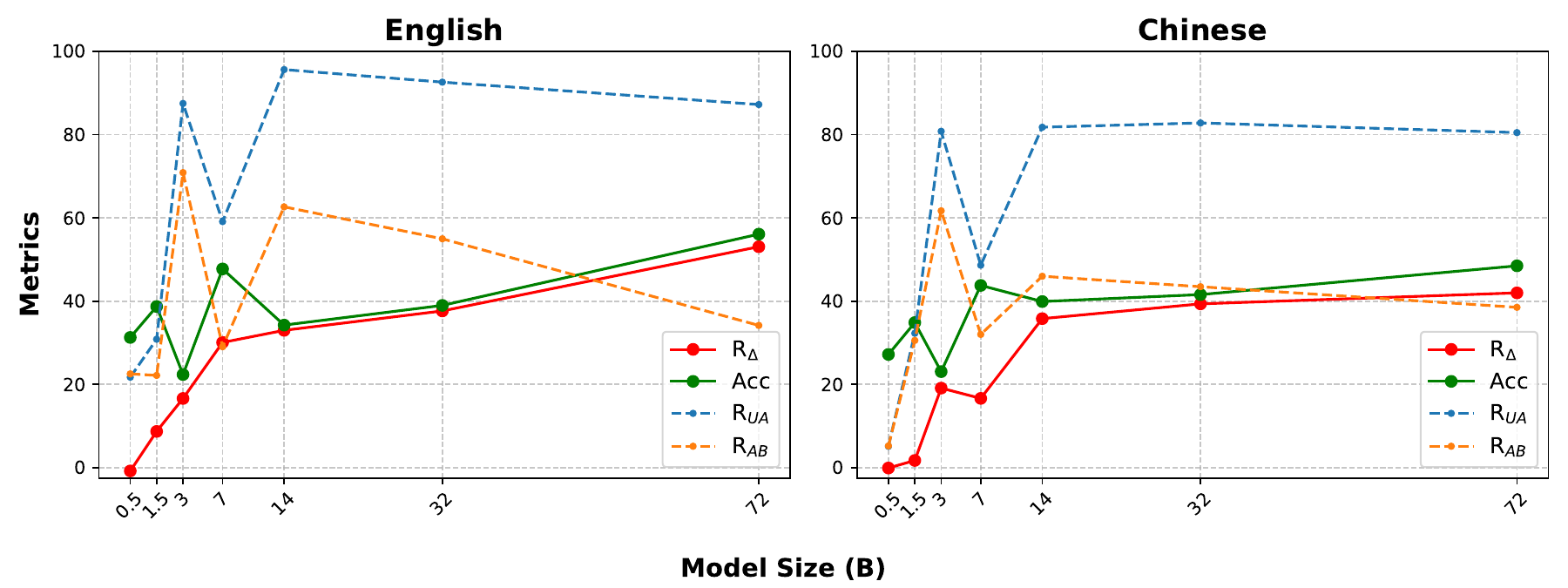}
    \caption{Refusal rate and Acc evaluated in Task 1 of Qwen2.5 series with parameters scaling from 0.5B to 72B. Detailed results are shown in Appendix~\ref{sec_app_task1_scaling}.}
    \label{fig_task1_qwen_multity}
\end{figure*}

\section{Experiments}
\label{sec_exp}
\subsection{Experiment Setup}

We conduct a sequence of experiments with various open-sourced and black-box LLMs, including Llama3 \citep{llama3modelcard}, Mistral0.2 \citep{jiang2023mistral}, Qwen2.5 \citep{qwen}, GLM4 \citep{zeng2022glm}, Gemini-1.5-pro \cite{gemini1.5_report}, and OpenAI series (GPT-4o-mini, GPT-4) \citep{achiam2023gpt}. Open-sourced LLMs we evaluated are corresponding \textit{Chat} or \textit{Instruct} versions around 7B. Our experiments are conducted based on the evaluation framework \textit{lm-evaluation-harness}~\citep{eval-harness} and more details can be found in Appendix \ref{sec_app_Experiment Setup}.

\subsubsection*{Metrics}
\noindent $\bullet$ Following \citet{liu2023prudent}, we obtain \textbf{\textit{refusal rate}} using lexical matching by identifying keywords that indicate denial, apology, or abstention. We conducted a human evaluation of sampled LLMs' responses, showing a strong alignment between the lexical matching results and human judgment. Details can be found in Appendix \ref{sec_app_human_evaluation_of_refusal_rate}. The refusal rates for UAQ and ABQ are denoted as \textbf{\textit{R$_{ua}$}} and \textbf{\textit{R$_{ab}$}}, respectively. 
We are also concerned with the difference between these two, denoted as \textbf{\textit{R$_\Delta$}}. The larger the \textbf{\textit{R$_\Delta$}}, the better LLM can discriminate UAQ and ABQ. 
Ideally, as the capacity of LLM enhances, \textbf{\textit{R$_{ua}$}} tends to 1, \textbf{\textit{R$_{ab}$}} tends to 0, and \textbf{\textit{R$_\Delta$}} tends to 1.

\noindent $\bullet$ For ABQ in UAQFact, we evaluate the accuracy (\textbf{\textit{Acc}}) of LLM's answer. We search LLM's responses by exact match based on the provided answer list.

\noindent $\bullet$ For Task 2, we first report the knowledge pass rate (\textit{\textbf{KPR}}), which measures the percentage of cases in which the LLM successfully passes the knowledge test. To enable a fair comparison of LLMs' ability to utilize internal knowledge across different KPR levels, we introduce a metric called knowledge-aware refusal‌ rate (\textit{\textbf{KRR}}):

\begin{equation}
\label{eq:KAR}
    \text{\textit{KRR}} = (1 + e^{-\text{\textit{R$_{\Delta}$}} \cdot {\text{\textit{KPR}}}^{-1}} )^{-1}
\end{equation}
A higher \textit{KRR} indicates better performance, with values ranging from 0 to 1.

\subsection{Task 1: Discriminating between UAQ and ABQ}
\label{sec_exp_task1}

In this task, we examine LLMs’ ability to discriminate UAQ and ABQ by directly providing UAQs/ABQs to LLMs, which is close to real-world application scenarios. Additionally, we analyze the relationship between LLMs' parameter size and their performance. The prompts we used are listed in Appendix \ref{sec_app_Evaluation}.

Results of Task 1 are listed in Table \ref{tab_task1_result}. All LLMs have a positive \textit{R$_\Delta$} in two languages. It indicates that LLMs have a certain ability to discriminate UAQ and ABQ when directly confronted with them. However, even the best LLM only achieves 63.26/38.40 \textit{R$_\Delta$} in English/Chinese, which means UAQFact can be a challenging benchmark for evaluating LLMs' ability to discriminate UAQ and ABQ. 

In English, black-box LLMs demonstrate superior \textit{R$_\Delta$} compared to open-sourced LLMs. While GPT-4o-mini shows a high refusal rate (85.05) for UAQs, it incorrectly refuses more ABQs than Mistral0.2, leading to lower \textit{Acc} scores (50.51 vs 54.32). Gemini-1.5-pro achieves the highest \textit{Acc} and maintains a \textit{R$_\Delta$} comparable to GPT-4, primarily due to its lower \textit{R$_{ab}$}.

We observe similar patterns in Chinese, though with a lower \textit{R$_\Delta$} compared to English. This suggests that Chinese questions pose greater challenges for LLMs in discriminating between UAQ and ABQ. We provide some cases in Appendix \ref{sec_app_task1_case_study}.

In summary, the above facts indicate that UAQFact is a very challenging benchmark for LLMs. 
For black-box LLMs, the \textit{R$_\Delta$} scores range from 42.91 to 63.26 in English and from 18.14 to 38.40 in Chinese, respectively.

\paragraph{\textbf{Model Scaling}} We report \textit{refusal rate} and \textit{Acc} of the Qwen2.5 series, including 7 versions with model scaling from 0.5B to 72B.
Figure \ref{fig_task1_qwen_multity} illustrates the trend of changes in metrics as model scale. Detailed results are listed in Appendix \ref{sec_app_task1_scaling}.
As LLMs' parameters scale up, there are noticeable increasing trends in \textit{R$_\Delta$}. This indicates that \textbf{larger LLMs can achieve better results in discriminating UAQ and ABQ}. 
On the other hand, we observe that \textit{R$_{ua}$} and \textit{R$_{ab}$} do not show the expected continuous increase or decrease. Instead, they exhibit consistent fluctuating patterns with each other. This suggests that the improvements LLMs achieved in \textit{R$_\Delta$} do not necessarily lead to better performance in both \textit{R$_{ua}$} and \textit{R$_{ab}$}: refusing more UAQs while refusing fewer ABQs.

\begin{table}[t]

\centering
\resizebox{0.48\textwidth}{!}{
\begin{tabular}{lcccccc}
\toprule
    \textbf{Language} & \multicolumn{3}{c}{\textbf{English}} & \multicolumn{3}{c}{\textbf{Chinese}} \\
    \cmidrule(lr){1-1} \cmidrule(lr){2-4} \cmidrule(lr){5-7}
    \textbf{Model} & \textbf{KPR $\uparrow$} & \textbf{R$_{\Delta}$ $\uparrow$} & \textbf{KRR $\uparrow$} & \textbf{KPR $\uparrow$} & \textbf{R$_{\Delta}$ $\uparrow$} & \textbf{KRR $\uparrow$}\\
    \midrule
    \multicolumn{7}{c}{\textbf{Open-sourced LLMs}}\\
    Llama3 & \textbf{73.11} & 20.89 & 57.10 & 52.71 & 9.95 & 54.71 \\
    Mistral0.2 & 68.92 & \textbf{30.88} & \textbf{61.02} & 40.21 & 7.92 & 54.91 \\
    Qwen2.5 & 70.57 & 30.07 & 60.49 & \textbf{60.44} & \textbf{16.63} & \textbf{56.84} \\
    GLM4 & 71.74 & 23.83 & 58.23 & 55.62 & 14.13 & 56.32 \\
    \midrule
    \multicolumn{7}{c}{\textbf{Black-box LLMs}}\\
    Gemini-1.5-pro & 69.03 & 54.24 & \textbf{68.69} & 76.74 & \textbf{38.40} & \textbf{62.26} \\
    GPT-4o-mini & 76.52 & 42.91 & 63.66 & 73.73 & 18.14 & 56.12 \\
    GPT-4 & \textbf{81.80} & \textbf{63.26} & 68.42 & \textbf{83.21} & 34.52 & 60.23 \\
    Average & 75.78 & 53.47 & 66.93 & 77.89 & 30.35 & 59.53 \\
    
\bottomrule
\end{tabular}
}
\caption{Performance of LLMs in Task 2 knowledge test. \textbf{KPR}: knowledge pass rate. \textbf{R$_{\Delta}$}: difference between R$_{ua}$ and R$_{ab}$ (from Table \ref{tab_task1_result}). \textbf{KRR}: knowledge-aware refusal rate. }
\label{tab_task2_KQ_acc_en}
\end{table}

\begin{figure*}[t]
    \centering
    \includegraphics[width=0.96\textwidth]{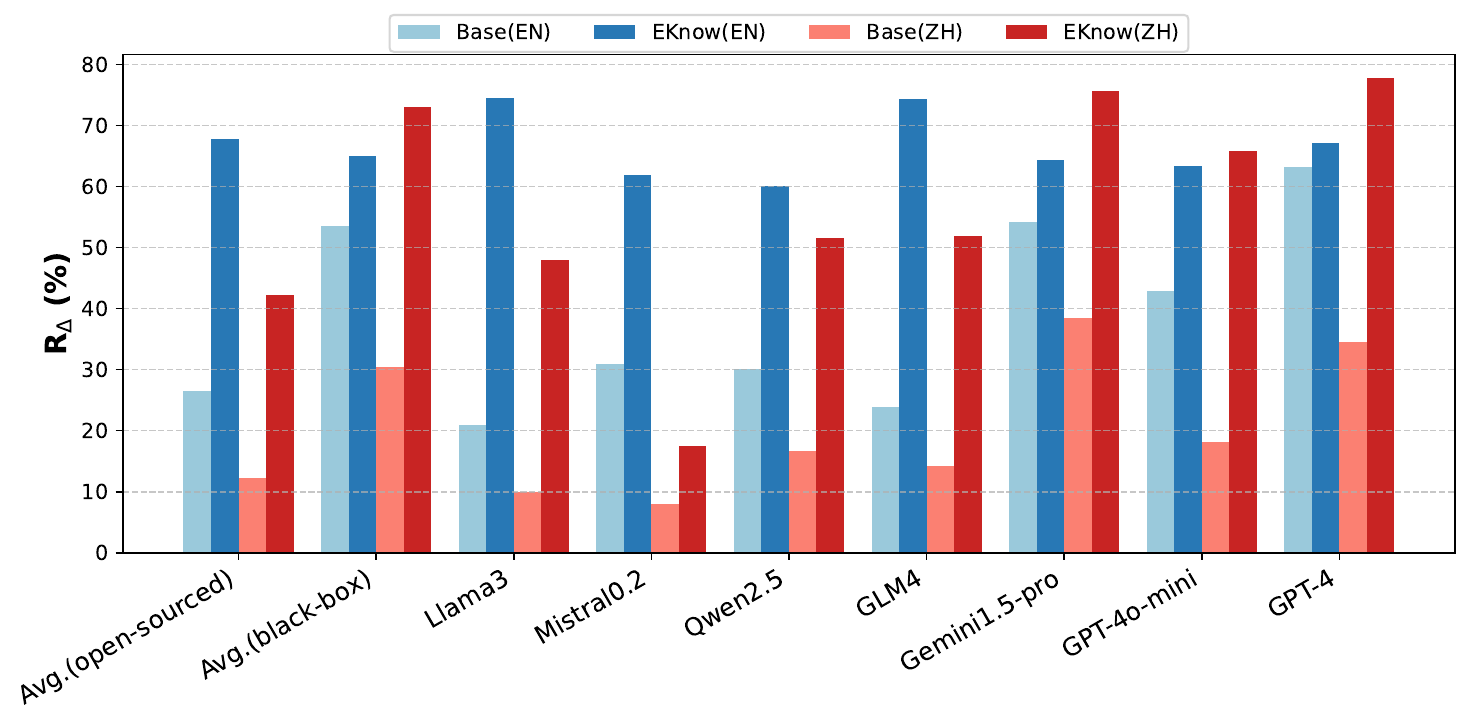}
    \caption{\textit{R$_{\Delta}$} Comparison Between Base and EKnow. \textit{EN} and \textit{ZH} are abbreviations for English and Chinese, respectively. Detailed results are shown in Appendix \ref{sec_app_task3}.}
    \label{fig_task3_CoTs_en}
\end{figure*}

\subsection{Task 2: Evaluating LLMs’ ability to utilize \myemph{internal} factual knowledge in handling UAQ}
\label{sec_exp_task2}
In this task, we provide LLMs with questions for knowledge testing and report their knowledge pass rate (\textit{KPR}) and knowledge-aware refusal rate (\textit{KRR}). Results are listed in Table \ref{tab_task2_KQ_acc_en}.

Black-box LLMs achieve higher average KPR than open-sourced LLMs, demonstrating stronger factual knowledge capability. Black-box LLMs also demonstrated a stronger ability in knowledge utilization compared with open-sourced LLMs on average. We observed that while Gemini-1.5-pro fails to achieve the highest KPR, it outperforms all other LLMs in terms of KRR, including GPT-4. Although Gemini 1.5 Pro's KPR in Chinese evaluation still shows a notable gap compared to GPT-4, its superior ability in internal knowledge utilization resulted in its R$_{\Delta}$ exceeding GPT-4 by 3.88.

Among open-source LLMs, Mistral0.2 excels in KPR for English, whereas Qwen2.5 excels in KPR for Chinese. Although the English KPR of Mistral0.2 is 1.65 points lower than Qwen2.5, it achieves a higher KRR, suggesting it applies its knowledge more efficiently despite having a smaller knowledge storage. However, Mistral0.2 exhibits a notable decline in KPR when processing Chinese inputs, resulting in a lower R$_{\Delta}$. These findings highlight that knowledge utilization and knowledge storage are critical determinants of LLMs' overall performance.

When evaluating the same set of Task 2 questions in Chinese versus English, most LLMs show decreased KPR, with Gemini-1.5-pro and GPT-4 being the exceptions, showing increases of 7.71 and 1.41 respectively. However, all models, including those with improved KPR, experience a decline in KRR when processing Chinese inputs, indicating greater difficulty in discriminating UAQ and ABQ in Chinese. Notably, Qwen2.5 and GLM4, despite their lower KPR compared to black-box LLMs, achieve slightly better KRR than GPT-4o-mini. It suggests that Qwen2.5 and GLM4 process a relatively stable knowledge utilization ability across languages. This can be attributed to their strategy during the training phase, where they carefully designed both the data composition and training task to enhance the model's multilingual capabilities.

In summary, the high KPR indicates that LLMs have stored extensive knowledge within their parameters. However, the comparatively low KRR reveals limitations in their ability to effectively utilize internal knowledge. 
Future research efforts should prioritize developing methods to enhance the utilization of LLMs' internal knowledge, thereby bridging the gap between knowledge storage and practical utilization.

\subsection{Task 3: Evaluating LLMs' ability to utilize \myemph{external} factual knowledge in handling UAQ}
\label{sec_exp_task3}

In this task, we provide questions and CoT with factual knowledge (\S~\ref{sec_data_task3}) to LLMs (EKnow), aiming to evaluate their ability to utilize external knowledge to correctly address UAQs. Figure~\ref{fig_task3_CoTs_en} shows the \textit{R$_\Delta$} of EKnow compared with Task 1 (Base). Detailed results are shown in Appendix~\ref{sec_app_task3}.~\footnote{In our preliminary experiments, we used EKnow provided in triples (Detailed in Appendix~\ref{app:task3_trp}.)}

With the help of CoT, which provides external reasoning clues with relevant factual knowledge, all LLMs demonstrate improved performance in Task 3 across both languages. Notably, open-sourced LLMs show more substantial gains compared to black-box LLMs, with their relative improvement (\textit{R$_\Delta$}) exceeding that of black-box LLMs. Llama3 exhibited the most remarkable enhancement, achieving the highest \textit{R$_\Delta$} with an improvement of 53.65 (20.89 vs 74.54). These results indicate Llama3's superior capability in leveraging external factual knowledge and effectively handling UAQ in English.

For black-box LLMs, external knowledge does help, but the impact is less significant compared to open-sourced LLMs in English. On one hand, black-box LLMs already achieve good performance by relying solely on their internal knowledge in Task 1, leaving limited room for further improvement. On the other hand, black-box LLMs show an overall decline in both R$_{ua}$ and R$_{ab}$. We find that black-box LLMs would refuse to respond at a certain rate due to uncertainty about UAQ-related information. With \textit{EKnow} in Task 3, black-box LLMs tend to provide more definitive responses. Some cases are listed in Appendix~\ref{sec_app_task3_case_study}.

In the Chinese setting, the results show different patterns. Black-box LLMs achieve greater improvements, with the average \textit{R$_\Delta$} in Chinese even surpassing that in English (73.08 vs 64.96). We also notice that the gap in average \textit{R$_\Delta$} between Chinese and English for black-box LLMs in Task 3 is smaller compared to Task 1 (8.12 vs 23.12). This suggests that black-box LLMs demonstrate more balanced capabilities across languages when leveraging external knowledge for UAQ. In contrast, the performance gap widened for open-sourced LLMs (25.49 vs 14.26), indicating that open-source LLMs still have room for improvement in utilizing cross-lingual external knowledge.

In summary, LLMs' ability to effectively utilize external knowledge remains a significant challenge. Even when provided with verified factual knowledge, the best \textit{R$_\Delta$} only reaches around 75\% in English. The performance tends to decline when using automatic retrieval methods. How to make full use of external knowledge can be an interesting research topic in future work.

\section{Conclusion}
This paper presents a bilingual dataset UAQFact for deeply evaluating the ability of LLMs to handle unanswerable questions.
With auxiliary factual knowledge, UAQFact can support two new tasks other than the classification task. These new tasks can comprehensively assess LLMs’ ability to utilize internal and external factual knowledge in handling UAQ, respectively. 

Our evaluation results indicate several promising research directions:
(1) Enhancing internal knowledge utilization: developing improved methods to activate and utilize factual knowledge already embedded within LLMs; 
(2) Strengthening external knowledge integration: advancing approaches to better incorporate external knowledge.

\section*{Limitations}
 Following \citet{liu2023prudent}, we use lexical matching to derive the metric \textit{refusal rate}. While human evaluations have confirmed the effectiveness of lexical matching, as evidenced by a Cohen's Kappa of 94.90 (Appendix \ref{sec_app_human_evaluation_of_refusal_rate}), there might remain a discrepancy between human evaluation and lexical matching results. Therefore, it is necessary to develop a more precise automated evaluation method in the future.

\section*{Acknowledgments}
This work is supported by the National Natural Science Foundation of China (Grant No. 62261160648, 62376177) and Provincial Key Laboratory for Computer Information Processing Technology, Soochow University. This work is also supported by Collaborative Innovation Center of Novel Software Technology and Industrialization, the Priority Academic Program Development of Jiangsu Higher Education Institutions. We would also like to thank the anonymous reviewers for their insightful and valuable comments.

\bibliography{custom}

\appendix

\section{Details for Factual Triple Sampling}
\label{sec_app_Factual Triple Sampling}

In this section, we provide examples of the factual triple sampling for three QTypes. The templates of SPARQL statements we used to query Wikidata are listed in Table \ref{tab_sparql}.

Denote the property list we obtained in \S~\ref{sec_question_generation} as \textbf{\textit{P$_l$}}, we illustrate the factual triple sampling procedure of three QTypes as follows (take the English version as an example): 

\paragraph{Inter}
We randomly sample two properties (\textit{editor}, \textit{cast member}) from \textbf{\textit{P$_l$}} and query with them: \textit{Join((?e$_1$, editor, ?ans), (?e$_2$, cast member, ?ans))} (\textit{Select Intersection} in Tabel \ref{tab_sparql}). 
If the query result exists, e.g. "\textit{[Die Rote Fahne (?e$_1$), The Eternal Jew (?e$_2$), \{Rosa Luxemburg, ...\} (?ans, a non-empty set)]}", we preserve "\textit{[(Die Rote Fahne, editor, ans$_1$) \& (The Eternal Jew, cast member, ans$_2$]}" as an instance of factual triple and "\textit{\{Rosa Luxemburg, ...\}}" as the planned answer for ABQ.
The \textit{ans$_1$} and \textit{ans$_2$} are full answer set fetched by querying \textit{(Die Rote Fahne, editor, ?ans$_1$)} and \textit{(The Eternal Jew, cast member, ?ans$_2$)} respectively (\textit{Select Tail} in Tabel \ref{tab_sparql}). 
For UAQ, we continue to query with the above properties separately, e.g. query \textit{(?e$_1$, editor, ?ans$_1$)} and \textit{(?e$_2$, cast member, ?ans$_2$)} by \textit{Select Factual Triples} in Table \ref{tab_sparql}.
Then we combine triples with no intersection of \textit{?ans$_1$} and \textit{?ans$_2$} as factual triples for UAQ, e.g. "\textit{[Enneads (?e1), editor, \{Porphyry, ...x$_i$\} (?ans$_1$, a non-empty set)] \& [The Sixth Sense (?e$_2$), cast member, \{Mischa Barton, ...y$_j$\} (?ans$_2$, a non-empty set, have no intersection with ?ans$_1$)]}". The planned answer to it is "\textit{None}", signifying that it possesses no answer.

\paragraph{Time}
We randomly select a property (\textit{spouse}) from \textbf{\textit{P$_l$}} and utilize the \textit{Select Time-related Information} query specified in Table \ref{tab_sparql} to retrieve the relevant factual triples. e.g. \textit{[Queen Paola of Belgium,spouse,?ans]} ,\textit{[?ans,start time,?time ]}. We choose the queried time \textit{?time} as the time constraint for the main question, and the \textit{?ans} as the answer of the ABQ. For UAQ, We randomly sample time points that can not fit the time constraint above, e.g. \textit{[Queen Paola of Belgium, spouse, ?ans]}, \textit{[?and, start time, ?sample time]}.
Then by querying the triples with the time, if we obtain an empty result set, we designate this sample time as the time constraint for the UAQ.

\paragraph{Dilemma}
We choose a property from \textbf{\textit{P$_l$}} and use the \textit{select factual triples} query in Table \ref{tab_sparql} to select factual triples, e.g. \textit{[Russell Banks, personal library at, ?ans]}. Then we query answer set \textit{ans} with \textit{Join(Russell Banks, personal library at, ?ans)} by \textit{Select Tail} in Table \ref{tab_sparql}. We randomly choose an answer from the answer set \textit{ans} and use the \textit{Select Options} in Table \ref{tab_sparql} to select and decide the corresponding planned options. 
All candidates in the planned options are incorrect for UAQ, and for ABQ, one candidate in the planned options is correct.

\section{Prompt}

\paragraph{Prompt for Template Generation} Here is the prompt we used for template generation.
\label{sec_app_Template_Generation}
\begin{tcolorbox}[title=Prompt for Template Generation]
\small
Turn the Property into a question template.
Take its Description in WikiData as a reference.
Return in the following format \{"Template":" "\}.
Some examples are shown below.

...

Property: [E1], editor, [Ans]

Description: editor is the person who checks and corrects a work (such as a book ...)

Template: Who is the editor of [E1]?

...

Property: \{\textit{property}\}

Description: \{\textit{description}\}

Template: 

=================================

\begin{CJK}{UTF8}{gbsn}
将关系转换为问句模板。请以WikiData中对关系的描述作为参考。以\{"Template":" "\}形式返回。下面是一些示例。

...

关系: [E1], 编辑者, [Ans]

描述: 编辑工作的编辑，如书或定期刊物...

模板: 谁是[E1]的编辑？

...

关系: \{\textit{property}\}

描述: \{\textit{description}\}

模板: 

\end{CJK}
\end{tcolorbox}

\paragraph{Prompt for Evaluation} are show in Table \ref{tab_eval_prompt}, which are used in \S \ref{sec_exp} .
\label{sec_app_Evaluation}

\begin{table*}[t]
    \small
    \centering
    \begin{tabularx}{0.92\textwidth}{lX}
    \toprule
        \textbf{Setting} & \textbf{Prompt}\\
    \midrule
        \multicolumn{2}{c}{\textbf{English}} \\
        \multirow{3}{*}{Task1} & Answer the given question in no more than one sentence. Please keep your answer short and concise. Return \#\#None\#\# if there is no suitable answer. \\
        & Q: \{\textit{question}\}. \\
        & A: \\ 
    \hdashline
        \multirow{4}{*}{Task 2} & The following are multiple choice questions (with answers) \\
        & Q: \{\textit{question}\} \\
        & (A) \{\textit{option[0]}\} (B) \{\textit{option[1]}\} (C) \{\textit{option[2]}\} (D) \{\textit{option[3]}\} \\
        & A: \\
    \hdashline
        \multirow{3}{*}{Task 3} & Answer the given question in no more than one sentence. Please keep your answer short and concise. Return \#\#None\#\# if there is no suitable answer. \\
        & Q: \{\textit{question}\}. \\
        & A: \{\textit{clue}\} \\ 
    \midrule
        \multicolumn{2}{c}{\textbf{Chinese}} \\
        \multirow{3}{*}{Task1} & \begin{CJK}{UTF8}{gbsn}在1句话以内回答给定问题。确保你的答案简短并简洁。如果你认为不存在合适答案，返回\#\#None\#\#。\end{CJK} \\
        & \begin{CJK}{UTF8}{gbsn}问题: \{\textit{question}\}.\end{CJK} \\
        & \begin{CJK}{UTF8}{gbsn}答案: \end{CJK} \\ 
    \hdashline
        \multirow{4}{*}{Task2} & \begin{CJK}{UTF8}{gbsn}下列是多项选择题（有答案）\end{CJK} \\
        & \begin{CJK}{UTF8}{gbsn}问题: \{\textit{question}\}.\end{CJK} \\
        & (A) \{\textit{option[0]}\} (B) \{\textit{option[1]}\} (C) \{\textit{option[2]}\} (D) \{\textit{option[3]}\} \\
        & \begin{CJK}{UTF8}{gbsn}答案: \end{CJK} \\ 
    \hdashline\\
        \multirow{3}{*}{Task 3} & \begin{CJK}{UTF8}{gbsn}在1句话以内回答给定问题。确保你的答案简短并简洁。如果你认为不存在合适答案，返回\#\#None\#\#。\end{CJK} \\
        & \begin{CJK}{UTF8}{gbsn}问题: \{\textit{question}\}.\end{CJK} \\
        & \begin{CJK}{UTF8}{gbsn}答案: \end{CJK} \{\textit{clue}\} \\ 
    \bottomrule
    \end{tabularx}
    \caption{Prompts for evaluation in \S \ref{sec_exp}.}
    \label{tab_eval_prompt}
\end{table*}

\section{Examples of Data in Three Tasks}

\paragraph{Example of Task1}
\label{sec_app_task1_example}

Table \ref{tab:main_que_examples_zh} shows examples of Task1 questions in Chinese.

\begin{table*}[t]
    \footnotesize
    \centering
    \resizebox{0.92\textwidth}{!}{
    \begin{tabularx}{\textwidth}{llX}
    \toprule
        QType & Description & Example\\
    \midrule
        Inter & Return intersection of two sets & (UAQ) \textbf{\textit{Q$_{i_1}$}}: \begin{CJK}{UTF8}{gbsn}谁既是九章集的编辑，同时也是第六感的演员？\end{CJK} \\ 
    \cmidrule(lr){3-3}
        & & (ABQ) \textbf{\textit{Q$_{i_2}$}}: \begin{CJK}{UTF8}{gbsn}\textit{谁既是红旗报的编辑，同时也是永远的犹太人的演员}？\end{CJK}\\ 
    \midrule
        Time & Consider time constraints & (UAQ) \textbf{\textit{Q$_{t_1}$}}: \begin{CJK}{UTF8}{gbsn}1957-1962年期间，爱尔福特的姊妹城市是哪个？\end{CJK}\\
    \cmidrule(lr){3-3}
        & & (ABQ) \textbf{\textit{Q$_{t_2}$}}: \begin{CJK}{UTF8}{gbsn}1971-1976年期间，爱尔福特的姊妹城市是哪个？\end{CJK} \\ 
    \midrule
        Dilemma & Provide candidate answer & (UAQ) \textbf{\textit{Q$_{d_1}$}}: \begin{CJK}{UTF8}{gbsn} 桑格斯是哪个部落的成员，莫霍克人还是康迪人？ \end{CJK}\\
    \cmidrule(lr){3-3}
        & & (ABQ) \textbf{\textit{Q$_{d_2}$}}: \begin{CJK}{UTF8}{gbsn} 桑格斯是哪个部落的成员，谢鲁斯克还是康迪人？\end{CJK}\\
    \bottomrule
    \end{tabularx}
    }
    \caption{Example of Chinese questions.}
    \label{tab:main_que_examples_zh}
\end{table*}

\paragraph{Example of Task2}
\label{sec_app_task2_example}

An example of Task 2 evaluation setting is shown as follows:
\begin{tcolorbox}
\small

\noindent \textbf{Task1 UAQ}: Who has been one of the head of government of Russian Soviet Federative Socialist Republic and also the father of Ramdas Gandhi?\\

\noindent \textbf{Task2 Questions} (asking UAQ relevant internal knowledge). The gold answer is shown in \textbf{bold}:

\noindent \textbf{Q1}: Who has been the head of government of Russian Soviet Federative Socialist Republic?

\noindent (A) Abdul Kahar of Brunei (B) Mahatma Gandhi \textbf{(C) Boris Yeltsin} (D) Salman Khan

\noindent \textbf{Q2}: The father of Ramdas Gandhi is?

\noindent (A) Boris Yeltsin (B) Syn (C) Lucius Tarquinius Collatinus \textbf{(D) Mahatma Gandhi}\\

\noindent \textbf{KPR=1}: Pass knowledge tests, e.g. choosing \textbf{(C)} for Q1 and \textbf{(D)} for Q2;

\noindent \textbf{KPR=0}: Fail at least one knowledge test, e.g. choosing (A)/(B)/(D) for Q1 or choosing (A)/(B)/(C) for Q2;

\end{tcolorbox} 

\paragraph{Example of Task3}
\label{sec_app_Reasoning_Clues}
Examples of reasoning clues we construct in \S~\ref{sec_data_task3} are shown in Table \ref{tab_app_Knowledge_Clues_example}, including 3 versions for 3 QTypes. These reasoning clues are used as external knowledge in Task 3.

\section{Manual Inspection of UAQFact}
\label{sec_app_manual_inspection}

We perform manual inspection during and after data construction:

1) During data construction, we conduct a comprehensive manual inspection of \textbf{all} question templates generated by GPT-3.5. Three annotators independently review each template. Any template marked as incorrect by one annotator is either discarded or revised. Initially, GPT-3.5 generated 1,259 question templates. After our rigorous inspection process, 395 templates are discarded, leaving 864 valid templates. Among these remaining templates, 229 templates are revised. Table \ref{tab:tmp_Question_Generation} presents examples of incorrect question templates originally generated by GPT-3.5 alongside their human-revised versions.

2) After data construction, we conducted a quality assessment by manually examining 900 randomly sampled questions from UAQFact. Our inspection confirms that all question templates are accurately aligned with their designated properties and question types. Additionally, we cross-validated the annotated factual knowledge against Wikipedia and Google to ensure accuracy.
In our analysis of 900 questions, we find that only a minimal portion (7 questions, 0.8\%) are incorrectly labeled as UAQs due to knowledge error in Wikidata. The vast majority of questions (893, 99.2\%) successfully passed our rigorous verification process.

\section{Experiment Setup}
\label{sec_app_Experiment Setup}
Open-sourced LLMs we evaluate are listed in Table \ref{tab:tab_model_cards}.
For all 7B LLMs, we set the \textit{temperature} to 0 and inference on V100 with \textit{dtype} set to \textit{float16}. For Qwen series LLMs, we infer those versions smaller than 32B locally following the above setting. For 72B, we fetch the response from API \footnote{https://api.together.ai/}.

For Task 1 and Task 3, we set \textit{output\_type} = \textit{generate\_until} and calculate the refusal rate by a lexical matching function defined by us. For Task 2, we set \textit{output\_type} = \textit{multiple\_choice} and use \textit{acc} as the metric, which is defined by \textit{lm-evaluation-harness}.

\begin{table}[t]
    \small
    \centering
    \resizebox{0.46\textwidth}{!}{
    \begin{tabular}{ll}
    \toprule
        \textbf{Model Name} & \textbf{Model Card in HF} \\ 
    \midrule
        Llama3 & meta-llama/Meta-Llama-3-8B-Instruct \\
        Mistral0.2 & mistralai/Mistral-7B-Instruct-v0.2 \\
        Qwen2.5 & Qwen/Qwen2.5-7B-Instruct \\
        GLM4 & THUDM/glm-4-9b-chat-hf \\
    \midrule
        \multicolumn{2}{l}{Qwen2.5 Series}\\
    \midrule
        0.5B & Qwen/Qwen2.5-0.5B-Instruct\\
        1.5B & Qwen/Qwen2.5-1.5B-Instruct\\
        3B & Qwen/Qwen2.5-3B-Instruct\\
        7B & Qwen/Qwen2.5-7B-Instruct\\
        14B & Qwen/Qwen2.5-14B-Instruct\\
        32B & Qwen/Qwen2.5-32B-Instruct\\
        72B & Qwen/Qwen2.5-72B-Instruct\\

    \bottomrule
    \end{tabular}
    }
    \caption{Open-sourced LLMs we evaluate and their corresponding model cards in Hugging Face. }
    \label{tab:tab_model_cards}
\end{table}

\begin{table}[t]
    \small
    \centering
    \resizebox{0.46\textwidth}{!}{
    \begin{tabular}{lccc}
    \toprule
         \textbf{Model} & \textbf{Lexical Matching} & \textbf{Human} & \textbf{Cohen’s Kappa}\\
         \midrule
         Llama3 & 35.33 & 33.22 & 94.11 \\
         Mistral0.2 & 47.33 & 41.22 & 89.42\\
         Qwen2.5 & 38.89 & 38.89 & 96.30\\      
         GLM4 & 57.44 & 57.33 & 99.77 \\      
         \midrule
         Average & 44.75 & 42.67 & 94.90\\               
    \bottomrule
    \end{tabular}
    }
    \caption{Refusal Rate obtained by lexical matching and human judgment on sampled data. We apply stratified random sampling to each LLM, drawing a sample of 900 cases based on Qtype and label ("UAQ"/"ABQ"), totaling 3,600 cases. Cohen’s Kappa shows a strong alignment between them.}
    \label{tab:human_label}
\end{table}

\begin{table}[t]
    \centering
    \small
    \resizebox{0.46\textwidth}{!}{
    \begin{tabularx}{\linewidth}{ll}
    \toprule
        \textbf{Type} & \textbf{Examples}\\
    \midrule
        Semantic & \redcross What is the function of [E1]'s GPU?\\
        Error & \greentick What type of GPU does [E1] use? \\
    \hdashline
        Missing  & \redcross What type of goods do shops typically sell?\\
        Slot & \greentick What type of goods do [E1] typically sell?\\
    \bottomrule
    \end{tabularx}
    }
    \caption{Incorrect question templates generated by GPT-3.5 and templates after human revision.}
    \label{tab:tmp_Question_Generation}
\end{table}

\begin{table}[t]
\small
\centering
\resizebox{0.46\textwidth}{!}{
\begin{tabular}{lcccccccc}
\toprule
    \textbf{Language} & \multicolumn{4}{c}{\textbf{English}} & \multicolumn{4}{c}{\textbf{Chinese}} \\
    \cmidrule(lr){2-5} \cmidrule(lr){6-9}
    \textbf{Metric} & \multicolumn{3}{c}{\textbf{Refusal Rate}} & \textbf{Acc} & \multicolumn{3}{c}{\textbf{Refusal Rate}} & \textbf{Acc}\\
    \cmidrule(lr){1-1} \cmidrule(lr){2-4} \cmidrule(lr){5-5} \cmidrule(lr){6-8} \cmidrule(lr){9-9}
    \textbf{Model Size} & \textbf{R$_{ua}$ $\uparrow$} & \textbf{R$_{ab}$ $\downarrow$} & \textbf{R$_\Delta$ $\uparrow$} & \textbf{Acc $\uparrow$} & \textbf{R$_{ua}$ $\uparrow$} & \textbf{R$_{ab}$ $\downarrow$} & \textbf{R$_\Delta$ $\uparrow$} & \textbf{Acc $\uparrow$}\\

    \midrule
    0.5B & 21.72 & 22.52 & -0.80 & 31.27 & 5.11 & \textbf{5.21} & -0.10 & 27.19 \\
    1.5B & 30.84 & \textbf{22.15} & 8.69 & 38.70 & 32.27 & 30.55 & 1.72 & 34.85 \\
    3B & 87.49 & 70.88 & 16.61 & 22.36 & 80.80 & 61.70 & 19.10 & 23.06 \\
    7B & 59.10 & 29.03 & 30.07 & 47.73 & 48.63 & 32.00 & 16.63 & 43.77 \\ 
    14B & \textbf{95.63} & 62.65 & 32.98 & 34.24 & 81.78 & 45.98 & 35.79 & 39.90 \\
    32B & 92.61 & 54.97 & 37.64 & 38.97 & \textbf{82.81} & 43.46 & 39.34 & 41.57 \\
    72B & 87.23 & 34.16 & \textbf{53.07} & \textbf{56.09} & 80.49 & 38.51 & \textbf{41.98} & \textbf{48.48} \\
\bottomrule
\end{tabular}
}
\caption{Refusal rate and Acc evaluated in Task 1 of Qwen2.5 series (Detailed results of Figure \ref{fig_task1_qwen_multity}).}  
\label{tab_task1_qwen_en}
\end{table}

\begin{table}[t]
    \small
    \centering
    \resizebox{0.46\textwidth}{!}{
    \begin{tabular}{lcccccccc}
    \toprule

        \multicolumn{1}{l}{\textbf{Metric}} & \multicolumn{2}{c}{\textbf{R$_{ua}$ $\uparrow$}} & \multicolumn{2}{c}{\textbf{R$_{ab}$ $\downarrow$}} & \multicolumn{2}{c}{\textbf{R$_\Delta$ $\uparrow$}} & \multicolumn{2}{c}{\textbf{Acc $\uparrow$}}\\
        \cmidrule(lr){1-1} \cmidrule(lr){2-3} \cmidrule(lr){4-5} \cmidrule(lr){6-7} \cmidrule(lr){8-9}
        \textbf{Model} & \textbf{Base} & \textbf{EKnow} & \textbf{Base} & \textbf{EKnow} & \textbf{Base} & \textbf{EKnow} & \textbf{Base} & \textbf{EKnow}\\

    \midrule
    \textbf{English} & \multicolumn{8}{c}{\textbf{Open-sourced LLMs}} \\
        Llama3 & 38.80 & 87.27 & 17.91 & 12.73 & 20.89 & 74.54 & 53.21 & 73.06 \\
        Mistral0.2 & 62.15 & 86.37 & 31.27 & 24.48 & 30.88 & 61.89 & 54.31 & 76.85 \\
        Qwen2.5 & 59.10 & 86.96 & 29.03 & 26.90 & 30.07 & 60.06 & 47.73 & 69.06 \\
        GLM4 & 55.05 & 94.50 & 31.22 & 20.09 & 23.83 & 74.41 & 49.63 & 74.93 \\
        Avg & 53.78 & 88.78 & 27.36 & 21.05 & 26.42 & 67.73 & 51.22 & 73.48 \\
    \hdashline
    & \multicolumn{8}{c}{\textbf{Black-box LLMs}} \\
        Gemini-1.5-pro & 66.50 & 67.64 & 12.25 & 3.22 & 54.24 & 64.42 & 69.62 & 86.36 \\
        GPT-4o-mini & 85.05 & 83.69 & 42.15 & 20.34 & 42.91 & 63.35 & 50.51 & 74.46 \\
        GPT-4 & 85.70 & 75.06 & 22.43 & 7.95 & 63.26 & 67.12 & 66.79 & 83.71 \\
        Average & 79.08 & 75.46 & 25.61 & 10.50 & 53.47 & 64.96 & 62.31 & 81.51 \\
    \midrule
    \textbf{Chinese} & \multicolumn{8}{c}{\textbf{Open-sourced LLMs}} \\
        Llama3 & 23.65 & 78.35 & 13.70 & 30.37 & 9.95 & 47.98 & 35.88 & 50.28 \\
        Mistral0.2 & 49.49 & 91.41 & 41.57 & 73.87 & 7.92 & 17.54 & 19.31 & 6.00 \\
        Qwen2.5 & 48.63 & 73.16 & 32.00 & 21.57 & 16.63 & 51.59 & 43.77 & 69.76 \\
        GLM4 & 47.03 & 74.57 & 32.90 & 22.71 & 14.13 & 51.86 & 41.53 & 70.51 \\
        Avg & 42.2 & 79.37 & 30.04 & 37.13 & 12.16 & 42.24 & 35.12 & 49.14 \\
    \hdashline
    & \multicolumn{8}{c}{\textbf{Black-box LLMs}} \\
        Gemini-1.5-pro & 57.42 & 86.29 & 19.01 & 10.67 & 38.40 & 75.62 & 53.98 & 72.40 \\
        GPT-4o-mini & 45.84 & 95.29 & 27.70 & 29.42 & 18.14 & 65.87 & 47.02 & 66.89 \\
        GPT-4 & 63.85 & 83.58 & 29.33 & 5.82 & 34.52 & 77.76 & 51.02 & 81.36 \\
        Average & 55.70 & 88.39 & 25.35 & 15.30 & 30.35 & 73.08 & 50.67 & 73.55 \\
    \bottomrule
    \end{tabular}
    }
    \caption{Performance of LLMs with the help of external knowledge (Task3). \textbf{Base}: Task 1 setting; \textbf{EKnow}: external knowledge in Task 3. (Detailed results of Figure \ref{fig_task3_CoTs_en}).}
    \label{tab_task3_CoTs_en}
\end{table}

\section{Detailed Experiment Results}

\paragraph{Task1: Parameter Scaling}
\label{sec_app_task1_scaling}
We report detailed results of Task1 (Figure \ref{fig_task1_qwen_multity}) in Table \ref{tab_task1_qwen_en}.

\paragraph{Task3: Evaluating LLMs' ability to apply external factual knowledge in handling UAQ}
\label{sec_app_task3}

The \textit{refusal rate} and \textit{Acc} of LLMs with the help of external knowledge (EKnow) are shown in Table \ref{tab_task3_CoTs_en}.

\paragraph{Provide external factual knowledge in triple}
\label{app:task3_trp}
\begin{table}[t]
\centering
\resizebox{0.46\textwidth}{!}{
\begin{tabular}{lcccc}
\toprule
    \textbf{Metric} & \multicolumn{3}{c}{\textbf{Refusal Rate}} & \textbf{Acc}\\
    \cmidrule(lr){1-1} \cmidrule(lr){2-4} \cmidrule(lr){5-5}
    \textbf{Model} & \textbf{R$_{ua}$ $\uparrow$} & \textbf{R$_{ab}$ $\downarrow$} & \textbf{R$_\Delta$ $\uparrow$} & \textbf{Acc $\uparrow$} \\

    \midrule
    \textbf{Base}\\
    Llama3 & 38.80 & 17.91 & 20.89 & 53.21 \\
    Qwen2.5 & 59.10 & 29.03 & 30.07 & 47.73 \\
    GLM4 & 55.05 & 31.22 & 23.83 & 49.63 \\
    
    \midrule
    \textbf{EKnow (CoT)} \\
    Llama3 & 87.27 & 12.73 & 74.54 & 73.06 \\
    Qwen2.5 & 86.96 & 26.90 & 60.06 & 69.06 \\
    GLM4 & 94.50 & 20.09 & 74.41 & 74.93 \\
    
    \midrule
    \textbf{EKnow (Triple)} \\
    Llama3 & 58.44 & 9.38 & 49.06 & 88.90 \\
    Qwen2.5 & 58.98 & 7.67 & 51.31 & 95.00 \\
    GLM4 & 52.91 & 17.54 & 35.37 & 81.26\\

\bottomrule

\end{tabular}
}
\caption{Preliminary Experiment: External knowledge provided in the form of triples is denoted as EKnow (Triple). External knowledge in CoT used in Chapter 3 is referred to here as EKnow (CoT).}

\label{tab:task3_trp_res}

\end{table}
In our preliminary experiments, we use EKnow in the form of triples, referred to as EKnow (Triple). We compare this with CoT form, denoted as EKnow (CoT), as described in \S~\ref{sec_task_definition}. Detailed results are listed in Table~\ref{tab:task3_trp_res}. EKnow (Triple) achieves a higher \textit{Acc} on ABQ, with a significant improvement in \textit{R$_\Delta$} compared to the Base. However, its \textit{R$_\Delta$} is not as high as that of EKnow (CoT). Since this paper focuses on UAQ, where \textit{R$_\Delta$} is the core evaluation metric, we proceed with Task 3 using CoT as the form of external factual knowledge. Future researchers interested in this area may conduct tests using other settings, similar to our preliminary experiments.

\section{Case Study and Discussion}

\subsection{Human Evaluation of Refusal Rate}
\label{sec_app_human_evaluation_of_refusal_rate}
We apply stratified random sampling to the output of 4 LLMs in Task 1, drawing a sample of 900 cases based on Qtype and label ("UAQ"/"ABQ"), which cumulates to a total of 3,600 cases. After human annotation, we calculate Cohen’s Kappa coefficients between the refusal rate obtained by lexical matching and human judgment through the function \textit{sklearn.metrics.cohen\_kappa\_score}. Table \ref{tab:human_label} summarizes human evaluation results. Cohen’s Kappa coefficients show a strong alignment between the refusal rate obtained by lexical matching and human judgment. This implies that the refusal rate obtained by lexical matching serves as a dependable metric for evaluation.

\subsection{Cases of LLMs' outputs}

\subsubsection{Cases of Task 1 output}
\label{sec_app_task1_case_study}

In Table \ref{tab_app_task1_case_study}, we present the outputs of LLM under Chinese and English inputs. Note that the English input on the left and the Chinese input on the right are semantically equivalent. Cases show that it is harder for LLMs to handle inputs in Chinese. Take GPT-4 as an example, it (A) provides a wrong answer that does not fit the time constraint "From 1998 to 2003"; (B) was misled by wrong candidate answers provided by UAQ; (C) fails to consider the constraint, provides a correct answer to question "Who is the mother of Ptolemy XI Alexander II?", which does not fit the other constraint "has been one of the heads of government of Sicily".

\subsubsection{Cases of Task 3 output}
\label{sec_app_task3_case_study}

We present two cases in Table \ref{tab_app_task3_case_study}. 
In case (A), Gemini-1.5-pro provides an incorrect answer that is unrelated to the UAQ in Base. Even with external knowledge (EKnow), Gemini-1.5-pro still fails to refuse this UAQ. However, it shows some improvement by providing an answer that is at least relevant to the UAQ (correctly identifying "Wilhelm II" as the spouse of Princess Hermine Reuss of Greiz). 
In case (B), GPT-4 refuses UAQ in Base but fails to maintain this refusal in EKnow. When we presented GPT-4 with Task 2's question without providing options, it expressed uncertainty about the question. This suggests that GPT-4's refusal in Base stems from its uncertainty about UAQ rather than its judgment based on internal knowledge. In EKnow, GPT-4 provides an incorrect response because EKnow partially addresses some of its initial concerns, leading it to attempt an answer.

\begin{table*}[t]
\small
\centering
\resizebox{0.92\textwidth}{!}{
\begin{tabularx}{0.92\textwidth}{lXX}
\toprule
Task1 Input & English & Chinese \\
\midrule
    \textbf{(A)} UAQ & Federico Bonazzoli played for which team from 1998 to 2003? & \begin{CJK}{UTF8}{gbsn}1998年到2003年期间，费德里科·博纳佐利是哪个团队的成员？\end{CJK}\\ 
    \hdashline
    Qwen2.5 & \redcross US Cremona & \redcross \begin{CJK}{UTF8}{gbsn}国际足联裁判员\end{CJK} (FIFA Referees)\\
    GPT-4 & \greentick \#\#None\#\# & \redcross \begin{CJK}{UTF8}{gbsn}AC米兰\end{CJK} (A.C. Milan)\\
\midrule
    \textbf{(B)} UAQ & Where was Ahwak recorded, Trident Studios or Olympic Studios? & \begin{CJK}{UTF8}{gbsn}Ahwak在哪里录制的， 三叉戟工作室还是奥林匹克录音室 ？\end{CJK}\\ 
    \hdashline
    Qwen2.5 & \redcross Trident Studios & \redcross \begin{CJK}{UTF8}{gbsn}奥林匹克录音室\end{CJK} (Olympic Studios)\\
    GPT-4 & \greentick None & \redcross \begin{CJK}{UTF8}{gbsn}奥林匹克录音室\end{CJK} (Olympic Studios) \\
\midrule
    \textbf{(C)} UAQ & Who has been one of the head of government of Sicily and also the mother of Ptolemy XI Alexander II? & \begin{CJK}{UTF8}{gbsn}在曾经担任过西西里岛的行政首脑的人中，谁是托勒密十一世的母亲？\end{CJK} \\
    \hdashline
    Qwen2.5 & \greentick None & \redcross \begin{CJK}{UTF8}{gbsn}雷纳托·史济法尼\end{CJK} (Renato Schifani)\\
    GPT-4 & \greentick None & \redcross \begin{CJK}{UTF8}{gbsn}克莱奥帕特拉一世\end{CJK} (Cleopatra Selene of Syria) \\
    
\bottomrule
\end{tabularx}
}
\caption{Cases of LLMs' outputs in Task 1.}
\label{tab_app_task1_case_study}
\end{table*}

\begin{table*}[t]
\small
\centering
\resizebox{0.92\textwidth}{!}{
\begin{tabularx}{0.92\textwidth}{lX}
\toprule
Model & Example \\
\midrule
    (A) Gemini-1.5-pro & [Task 1 (UAQ)] Who has been the head of government of Goebbels cabinet and also the spouse of Princess Hermine Reuss of Greiz? \\
    Base & \redcross Adolf Hitler \\
    EKnow & \redcross Wilhelm II \\
\midrule
    (B) GPT-4 & [Task 1 (UAQ)] Who has been one of the head of government of Villavaliente and also the mother of Anne Frank?\\ 
    Base & \greentick None \\
    EKnow & \redcross Edith Frank-Holländer\\
    Response of Task 2 & This question cannot be answered without additional specific information as Villavaliente does not seem to refer to a known national or municipal government. \\
\bottomrule
\end{tabularx}
}
\caption{Cases of LLMs' outputs in Task 3.}
\label{tab_app_task3_case_study}
\end{table*}

\begin{table*}[t]
    \small
    \centering
    \resizebox{0.92\textwidth}{!}{
    \begin{tabularx}{\textwidth}{lX}
    \toprule
        \textbf{Function} & \textbf{SPARQL}\\
    \midrule
        \multirow{2}{*}{Select Property} & \textit{SELECT ?property ?propertyLabel ?propertyDescription WHERE \{ ?property a wikibase:Property . OPTIONAL \{ ?property skos:altLabel ?altLabel . FILTER (lang(?altLabel) = "en") \} SERVICE wikibase:label \{ bd:serviceParam wikibase:language "en" .\}\}} \\ 
         & \textit{SELECT ?property ?propertyLabel ?propertyDescription WHERE \{ ?property a wikibase:Property . OPTIONAL \{ ?property skos:altLabel ?altLabel . FILTER (lang(?altLabel) = "zh") \} SERVICE wikibase:label \{ bd:serviceParam wikibase:language "zh" .\}\}} \\ 
    \midrule
        \multirow{2}{*}{Select Property Description} & \textit{SELECT ?y WHERE \{wd:\%prop schema:description ?y. FILTER(LANG(?y) = 'en').\}}  \\
        & \textit{SELECT ?y WHERE \{wd:\%prop schema:description ?y. FILTER(LANG(?y) = 'zh').\}}\\
    \midrule
        Select Factual Triples & \textit{SELECT DISTINCT ?x ?y WHERE \{?x wdt:\%prop ?y .\} LIMIT 100} \\
    \midrule
        Select Options & \textit{SELECT ?y WHERE \{ wd:\%qid wdt:P31 ?x. ?y wdt:P31 ?x.\} LIMIT 100} \\
    \midrule
        Select Time-related Information & \textit{SELECT DISTINCT ?ans ?start  ?end ?point WHERE \{ wd:\%qid p:\%prop ?ans. OPTIONAL \{ ?ans pq:P580 ?start. \} OPTIONAL \{ ?ans pq:P582 ?end. \} OPTIONAL \{ ?ans pq:P585 ?point. \} FILTER((BOUND(?start)) || (BOUND(?end)) || (BOUND(?point))). \} LIMIT 20} \\
    \midrule
        Select Intersection & \textit{SELECT DISTINCT ?x ?y WHERE \{?x1 wdt:\%p1 ?y . ?x2 wdt:\%p2 ?y .\} LIMIT 100}\\
    \midrule
        Select Tail & \textit{SELECT DISTINCT ?y WHERE \{wd:\%x wdt:\%p1 ?y .\} LIMIT 100}\\
    \bottomrule
    \end{tabularx}
    }
    \caption{SPARQL templates for \S \ref{sec_factual_triple_sampling} Factual Triple Sampling.}
    \label{tab_sparql}
\end{table*}

\begin{table*}[t]
    \small
    \centering
    \resizebox{0.92\textwidth}{!}{
    \begin{tabularx}{\linewidth}{lX}
        \toprule
            \textbf{Version} & \textbf{Example} \\
        \midrule
            \textbf{English} &\\
        \midrule
            Inter & The question can be split into 2 sub-questions, denoted as q1 and q2. q1 is "\textbf{The editor of Enneads is?}?". q2 is "\textbf{The Sixth Sense's cast members are}?". The answer to q1 is "\textbf{Porphyry}". The answer to q2 is "\textbf{Mischa Barton, ...}". Combine the answers to q1 and q2 to make a judgment. If there is an intersection of the answers to the sub-questions, then the answer to the question is this intersection. If there is no intersection of the answers to the sub-questions, then there is no answer to the question. In summary, my answer is: \\
        \hdashline
            Time & This question can be split into the main question "Erfurt was twinned with which city?" and time "from 1957 to 1962". Through the auxiliary question, "When did Estadio GEBA start participating in association football the first time?", we obtain the START-TIME of the main question sentence "Erfurt was twinned with which city?", the answer of the auxiliary question is "1971". Determined whether "from 1957 to 1962" > START-TIME 1971. If the comparison condition is met, there is an answer, then the corresponding answer will be replied. If the comparison condition is not met, there is no answer to the question. In summary, my answer is: \\
        \hdashline
            Dilemma & This question is a dilemma. First we focus on the main problem, then the problem becomes: "What tribe does Segestes belong to, Mohawk people or Khamti people?", and the following options are "Mohawk people or Khamti people". The answer to the previous question is "Cherusci". If the answer appears in the two options, this is the answer, otherwise, there is no answer to the question. In summary, my answer is: \\
        \midrule
            \textbf{Chinese} &\\
        \midrule
            Inter & \begin{CJK}{UTF8}{gbsn}该问句可被拆分为2个子问题，记为q1和q2。q1是"\textbf{九章集的编辑是？}"。q2是"\textbf{第六感的演员有？}"。q1的答案是"\textbf{波菲利}"。q2的答案是"\textbf{美莎·芭顿, ...}"。结合q1和q2的答案进行判断。如果子问题答案有交集，则该问句的答案为此交集。如果子问题答案没有交集，则该问句没有答案。综上，我的答案是:\end{CJK}\\
        \hdashline
            Time & \begin{CJK}{UTF8}{gbsn}该问句是一个关于时间约束的问句，它可被拆分为主问句"爱尔福特的姊妹城市是哪个?"和时间"1957-1962年",通过辅助问句"爱尔福特和杰尔第一次成为友好城市的开始时间？"得到主问句的START\_TIME，这个辅助问句的答案是"1908",判断"1900年至1905年"是否与START\_TIME "1908"有交集，如果有交集则有答案，则回复对应的答案，如果没有交集，则该问句没有答案。综上，我的答案是:\end{CJK}\\
        \hdashline
            Dilemma & \begin{CJK}{UTF8}{gbsn}该问句是一个假两难问题。首先我们忽略候选选项，那么问题变为了："\textbf{桑格斯是哪个部落的成员， 莫霍克人还是康迪人？}"，问句中的两个选项是"莫霍克人还是康迪人"。这个问题的答案是"\textbf{谢鲁斯克}"，这些答案是否出现在了后续的选项中？如果出现了那么出现的便是答案，若没有出现则该问句没有答案。综上，我的答案是:\end{CJK} \\
        \bottomrule
    \end{tabularx}
    }
    \caption{Examples of reasoning clues (CoT) in Task 3.}
    \label{tab_app_Knowledge_Clues_example}
\end{table*}

\end{document}